\documentclass{article}
\usepackage{a4wide}
\usepackage{epsfig,latexsym}
\usepackage[utf8]{inputenc}
\usepackage{verbatim}
\usepackage{amsmath,amsfonts,amssymb}
\usepackage{multirow}
\usepackage{color}
\usepackage{algorithm}
\usepackage{algorithmic}
\usepackage{natbib}
\usepackage{graphicx}
\graphicspath{{}{../}{./}}
\usepackage[caption=false]{subfig}

\begin{document}
\begin{titlepage}
\begin{center}

  \begin{figure}[ht]
    \centering
    \includegraphics{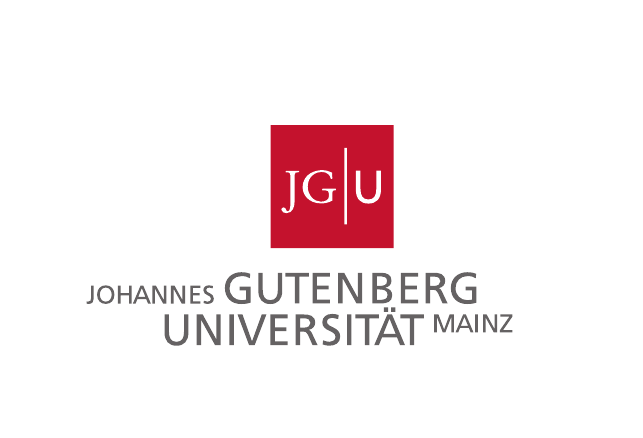}
  \end{figure}

\fbox{\parbox[b][7cm][c]{5in} { \fboxrule 2mm
\fboxsep 0.5cm 
\centering

{\large\bf Training a Restricted Boltzmann Machine for classification by labeling model samples}\\

\addvspace{0.2in}
{\bf Malte Probst, Franz Rothlauf}\\

\addvspace{0.3in}

Technical Report 04/2012 \\
April 2012}}

\vspace*{0.7in}

{ \large \bf Technical Report in Information Systems\\ and Business Administration}\\

\vspace*{1.6in}

 \line(1,0){380}

{{\bf Johannes Gutenberg-University Mainz}\\
Department of Information Systems and Business Administration\\
D-55128 Mainz/Germany\\
Phone +49 6131 39 22734, Fax +49 6131 39 22185\\
E-Mail: sekretariat[at]wi.bwl.uni-mainz.de\\
Internet: {\tt http://wi.bwl.uni-mainz.de}\\
}
\end{center}
\end{titlepage}

\title{Training a Restricted Boltzmann Machine for classification by labeling model samples}

\author{
\Large{Malte Probst \& Franz Rothlauf}\\
    probst/rothlauf@uni-mainz.de\\
       Department of Information Systems \& Business Administration\\
       Johannes-Gutenberg Universität Mainz, Germany
}
\date{}
\maketitle

\begin{abstract} 
We propose an alternative method for training a classification model. Using the MNIST set of handwritten digits and Restricted Boltzmann Machines, it is possible to reach a classification performance competitive to semi-supervised learning if we first train a model in an unsupervised fashion on unlabeled data only, and then manually add labels to model samples instead of training data samples with the help of a GUI. This approach can benefit from the fact that model samples can be presented to the human labeler in a video-like fashion, resulting in a higher number of labeled examples. Also, after some initial training, hard-to-classify examples can be distinguished from easy ones automatically, saving manual work.
\end{abstract} 

\section{Introduction}
\label{intro}
When solving classification problems in a supervised or semi-supervised fashion, it is always necessary to somehow label samples from the training data. Incorporating these labeled examples into the training process enables the model to assign a class label either directly to an unknown example or to the implicit category that the example belongs to. A common approach is applying labels to all or to a subset of the training examples prior to training. This process can be very time-consuming, especially if there are many examples and a large subset of them should be labeled. Using semi-supervised learning, it is possible to train a sufficient model while having only a subset of the training data enriched with labels \cite{Chapelle2010}. However, it is still necessary to label some samples prior to training.\\
This paper presents an alternative approach for the classification of images, which works in the reverse order: First, train a generative model of the data and afterwards apply labels to samples from the trained model. Similar ideas have been pursued in the field of face recognition, e.g. by \citet{Tian2010} using unsupervised clustering prior to a manual labeling task, however, we want take a more general approach. Reversing the order has some advantages over the classical way: First, it is possible to label more examples in a shorter period of time by showing the human labeler a constantly changing stream of model samples.
Second, it is possible to prevent the user from manually labeling examples similar to the ones that the model can already firmly classify. By trying to maximize the additional information in each new training example this aspect is similar to \textit{active learning}/\textit{selective sampling} proposed in \citet{Cohn1994}.\\
There are some caveats to this approach: If, at the time of the training, there is no label information, the parametrization of the training process must rely on metrics like the reconstruction error. Also, the samples generated by the model must be human-interpretable in order to perform the labeling.\\
Using the MNIST data set of handwritten digits, we show that the post-labeling approach is competitive to a semi-supervised training scenario.
\section{Preliminaries}
\subsection{Semi-supervised learning and generative models}
Semi-supervised training is a hybrid of supervised and unsupervised training \cite{Chapelle2010,zhu2009}. In unsupervised training settings, we try to find interesting structures in the a set $X$ consisting of $n$ training examples ${x_1,...,x_n}$ without explicitly assigning classes or labels to those structures, e.g. for clustering or statistical density estimation. In a supervised setting, we are interested in finding a mapping of a variable $x$ to another variable $y$ in a training set consisting of example pairs $(x_i,y_i)$, i.e. we are facing a classification task. In order to solve a supervised learning problem, it is possible to use discriminative or generative models. A discriminative model tries to directly learn the relationship between $x_i$ and $y_i$, often by trying to directly estimate the conditional probability $p(y|x)$ of a label given a data point. Using a generative model, the approach is more related to the unsupervised case: By learning the structure of the data in $X$, generative models try to estimate the class-conditional probability $p(x|y)$ or the joint probability $p(x,y)$ and retain the conditional probability $p(y|x)$ using Bayes' rule \cite{Chapelle2010}. Given a well-trained generative model, it is therefore often possible to draw samples from the model that resemble training data, as they come from the same probability distribution. Recently, a class of generative models called Restricted Boltzmann Machines has been widely used for discrimination tasks such as digit classification \cite{Hinton2006}, phone recognition \cite{Dahl2010} or document classification \cite{HintonSalakhutdinov2011}.
\subsection{Restricted Boltzmann Machines}
Restricted Boltzmann Machines are stochastic, energy-based neural network models \cite{Smolensky1986}. An RBM consists of a visible layer $v$ and a hidden layer $h$, connected by weights $w_{ij}$ from each visible neuron $v_i$ to each hidden neuron $h_j$, forming a bipartite graph.
They can be trained to model the joint distribution of the data, which is presented to the visible layer, and the hidden neurons by adjusting the weights $w_{ij}$ and biases $b_i$ and $c_i$. The neurons of the hidden layer are often referred to as \textit{feature detectors}, as they tend to model features and patterns occurring in the data, thus capturing the structure in the training data. The probability $P$ that a hidden neuron $h_j$ is active depends on the activation of the visible units $v_i$ and the bias of the  hidden neuron $c_j$, thus $P(h_j=1|v)=sigm(\sum_{i}w_{ij}v_i+c_j)$, with $sigm()$ being the logistic function $sigm(x)=\frac{1}{1+e^{-x}}$. The probability $P(v_i=1|h)$ that a visible unit $v_i$ is active given the hidden layer activations $h_j$ is, in turn, equal to $sigm(\sum_{j}w_{ij}h_j+b_i)$, with $b_i$ being the bias of neuron $v_i$. Calculating $P(h|v)$ and $P(v|h)$ is therefore easy and efficient .\footnote{Note that all neurons are modeled as binomial random variables, this can be generalized to any exponential family distribution, see e.g. \citet{Welling2005} or \citet{BengioTR2006}}\\
The energy function
\begin{center}$E(v,h)=-\sum_{i,j}v_ih_jw_{ij}-\sum_{i}v_ib_i-\sum_{j}h_jc_j$\end{center}
defined on the RBM associates a scalar energy value for each configuration of visible neurons $v$ and hidden neurons $h$. The probability $P(v,h)$ of a joint configuration is proportional to its energy:
\begin{center}$P(v,h)\propto e^{-E(v,h)}$\end{center}
It is now possible to marginalize over all hidden configurations to obtain the probability $P(v)$ of a visible vector (see \citet{Hinton2002} for details).
\subsection{Training RBMs using contrastive divergence}
To train an RBM on a data set, it is necessary to increase the probability (= lower the energy) of training data vectors and decrease the probability of configurations that do not resemble training data. This can be done by updating the weights $w_{ij}$ following the log likelihood gradient
$\frac{\partial log(P(v))}{\partial w_{ij}}$. It can be shown that this partial derivative is a sum of two terms usually referred to as the positive and negative gradient, which is why the training algorithm is called \textit{contrastive divergence} (CD) \cite{Hinton2002}. The resulting update rule for the weights is
\begin{center}$\delta w_{ij}\propto <v_i,h_j>^{data}-<v_i,h_j>^{model}$\end{center}
with $<v_i,h_j>$ being the expected value that $v_i$ and $h_j$ are active simultaneously. The first term (positive gradient) is calculated after initializing the visible layer with a data vector from the training set and subsequently activating $h$ given $v$. The second term (negative gradient) is calculated when the model is running freely, that is after a potentially infinite number of Gibbs sampling steps $v\rightarrow h\rightarrow v\rightarrow ...$. As the negative gradient is intractable, it is often approximated using only $N$ steps of sampling after initializing the visible neurons with data (CD-N). In practice, this approximation works pretty well (see e.g. \citet{Hinton2006}).\\
To learn a labeled data set, we simply extend the visible layer to also capture label data (e.g. a one-hot vector representing the label classes) and add an extra set of label weights $w^L_{kj}$ connecting the $k$ labels to the  hidden neurons. The learning rule for the label weights and biases remains unchanged.
\section{Post-labeling of MNIST digit model samples with an RBM}
\subsection{Overview}
\begin{figure}[ht]
\vskip 0.2in
\begin{center}
\centerline{\includegraphics[width=0.9\columnwidth]{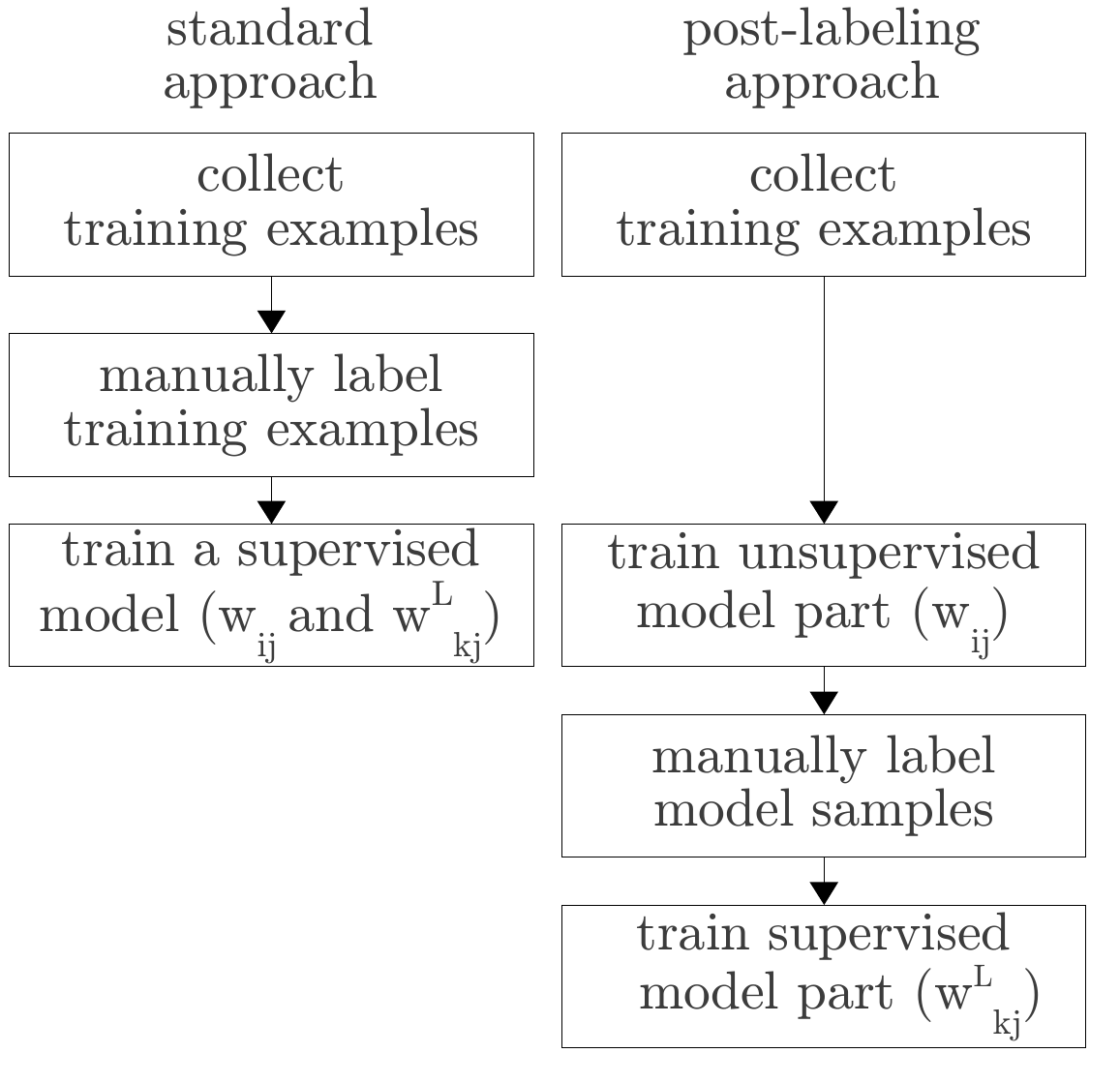}}
\caption{Blueprint of the required steps to train a classification RBM. Left-hand side: standard approach, right-hand side: post-labeling approach}
\label{model_ablauf}
\end{center}
\vskip -0.2in
\end{figure}

Figure \ref{model_ablauf} compares the steps of the standard approach to train a classification RBM with the post-labeling approach pursued in this paper. The standard approach first collects training data and then manually applies labels to the data, or to a subset of the data. Afterwards, a (semi-) supervised model is trained on labeled data, simultaneously learning both the regular weights $w_{ij}$, connecting the visible neurons to the features, and label weights $w^L_{kj}$, connecting the label neurons the features.\\
With post-labeling, we change the order: after collecting data, we train an RBM in an unsupervised fashion on the unlabeled data, thus only updating the regular weights $w_{ij}$. Afterwards, we let the model generate samples and apply labels to those samples. We then use the labeled samples to update the label weights $w^L_{kj}$ in a supervised way.
\subsection{Data set}
We used the MNIST database of handwritten digits for our experiments \cite{MNIST}. The data set contains 60,000 labeled training examples and 10,000 labeled test examples of 28*28 pixel images of handwritten digits in ten classes. When performing the semi-supervised or unsupervised learning tasks, we remove the labels.
\subsection{Models}
\label{models}
We perform the post-labeling tests on a Restricted Boltzmann Machine with 784 (=28*28) visible neurons $v_i$ and 225 hidden neurons $h_j$ (feature detectors). In order to validate the competitiveness of the post-labeling approach, we compare it to an RBM of the same size - with the visible layer extended by $k=10$ label neurons - trained on labeled data in a supervised (all data labeled) or semi-supervised (only a subset labeled) fashion. During the initial training, the post-labeling RBM thus only has one set of weights $w_{ij}$, whereas the classic RBM has a second set of label weights $w^L_{kj}$.\\
We train both models networks using the training algorithm CD-10 and 50,000 images from the training set. The remaining 10,000 examples are held out in order to find feasible parameters (such as the learning rate) for the supervised model.\footnote{We do not optimize the training procedure, as the resulting comparison is subjective, see section \ref{biases}} We use the reconstruction error (sum of squared pixel-wise differences between data and one-step reconstruction) to measure the training progress of the unsupervised model trained on data without labels. This is one of the main caveats of the post-labeling method: the reconstruction error can be misleading, especially when learning parameters are adapted during the training \cite{Hinton2010techreport}. Also, the reconstruction errors between different learning algorithms can differ without giving a proper hint to model quality.\footnote{During our experiments, the one-step reconstruction error on models trained with CD-1 is around 7, whereas the models trained with CD-10 shows a reconstruction error of 12. Nevertheless, the visual quality and in particular temporal stability of the representations on repeated Gibbs samplings is better with CD-10, which is also known to produce better results on discriminative tasks, given sufficient training time (see e.g. \citealt{Tieleman2008}).}
\subsection{Interactive post-labeling phase}
\begin{figure}[ht]
\vskip 0.2in
\begin{center}
\centerline{\includegraphics[width=\columnwidth]{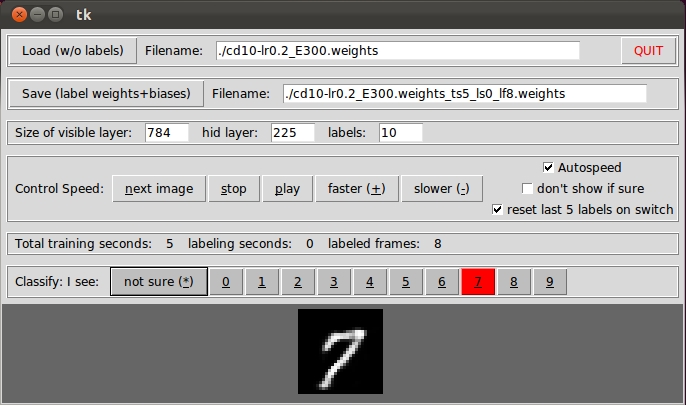}}
\caption{Screenshot of the labeltrainer GUI. The current sample from the model is displayed in the bottom, the highlighted button reflects the user-assigned label.}
\label{screenshot_labeltrainer_gui}
\end{center}
\vskip -0.2in
\end{figure}

\begin{figure}[ht]
\vskip 0.2in
\begin{center}
\centerline{\includegraphics[width=\columnwidth]{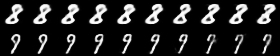}}
\caption{Sequences of generated samples from an RBM trained on unlabeled MNIST data. Between each two images, there is one step of Gibbs sampling ($v \rightarrow h \rightarrow v$). The first row shows a constantly changing eight (which might transition into a three in one of the subsequent images), the second row shows a transition from a nine to a seven.}
\label{samples_merged}
\end{center}
\vskip -0.2in
\end{figure}

The goal of the post-labeling phase is to find proper label weights $w^L_{kj}$. For this purpose, we developed a GUI that shows samples from the model to a human labeler, who can activate the corresponding class using the keyboard or mouse (see Fig. \ref{screenshot_labeltrainer_gui}). We initialize the visible layer with a randomly chosen (unlabeled) image from the training set and then let the model perform repeated Gibbs sampling between the visible and the hidden layer of the underlying RBM.
This leads to a slight deformation of the shown image in each sampling step, while the model traverses along a low-energy ravine in the energy landscape. If the model produces good reconstructions, the user observes slowly changing samples that belong to the same class, and potentially class transitions (see Figure \ref{samples_merged}). The displayed image is constantly updated at a speed of approx. 6 frames/second, which is adjustable in the GUI. The user's task is to activate the corresponding class as soon as the observed image firmly resembles one of the classes. The selected class label stays active until the user presses the "unsure" button or another class button. This leads to a high number of labeled samples, as the display resembles a video of "morphing" digits. After 30 Gibbs iterations, the visible neurons are initialized with the next random image from the training set.
\\
\subsection{Online learning while labeling}
There are two possibilities for training the label weights $w^L_{kj}$. The first is to perform online learning during the post-labeling phase. Whenever a label is activated by the user, we update the label weights proportional to an approximation of the positive and the negative gradient at the same time. In this case, the positive gradient is $<a_k,h_j>$, with $a_k$ being the $k$'th class label activation (as given by the user) and $h_j$ the probability of the $j$'th feature being active. The negative gradient approximation is $<l_k,h_j>$ with $l_k$ being the probability of the $k$'th label being active (as reconstructed by $h$). Thus, we strengthen connections from active features to the correct label and penalize connections from active features to the potentially wrong, reconstructed label. The biases are updated accordingly. We activate online learning in the GUI by default.
\subsection{Offline learning after labeling}
Alternatively, it is possible to train the label weights $w^L_{kj}$ in an offline fashion after the manual labeling of model samples in the GUI. We save all frames labeled in the GUI and used them to train the label weights $w^L_{kj}$ using standard CD-1. Again the update for the label weights is proportional to $<a_k,h_j>^{data}-<l_k,h_j>^{model}$.  The only difference to the online learning is that we can cycle through the labeled training set multiple times, thus the negative gradient may change during the course of the training, resulting in a better approximation. The weights $w_{ij}$ remain unchanged during the learning phase.
\subsection{Improvements and Tweaks}
It is possible to improve the ease of use of the labeling GUI and the resulting labeling quality using a few tweaks. First, we can automatically control the speed of the image stream that is presented the user. After a few minutes of training, the model already assings a reasonably high probability to the correct class for "common" samples (online learning is activated). On the contrary, if the current sample is visually distant from the previously labeled samples, the model doesn't assign a high probability to any label - it is unsure which label to pick for this example. Thus, it is possible to decrease the display speed for samples that seem unknown, thus allowing the user to make a more precise pick of the label (especially on class transitions). We implemented this tweak in the GUI as "autospeed" and activated it by default (see Fig. \ref{screenshot_labeltrainer_gui}).\\
Analogously, it is possible to bias the choice of samples from the training set to initialize the image (active sampling). If the probability for a label is very high ($>$80\%) the GUI can directly skip the example and try the next one. Although this approach channels the user's attention to samples where the model is still unsure, it deprives the learning process of the chance to detect confident misclassifications. Thus this technique shouldn't be used right away but only after some training. We implemented this "don't show if sure" concept in the GUI and asked users to activate it after the first five minutes of training.\\
We also added the possibility to automatically undo the last five update steps if  the user changes his opinion on a displayed image (class changes and changes from a class to \textit{unsure}). Initial tests showed that when running on higher speeds, the reaction time of a user usually allows some wrong labels to slip in in case of a class transition or image degradation.\\  
If the reconstructions of the model are too stable to produce a constantly changing stream, it is possible to implement a set of "fast weights" as in \citet{Tieleman2009}. Those fast weights can add a temporary penalty to the areas of low energy just visited, thus forcing the model to wander around. We didn't implement this tweak as of now.
\subsection{Results}
\begin{figure}[ht]
\vskip 0.2in
\begin{center}
\centerline{\includegraphics[width=\columnwidth]{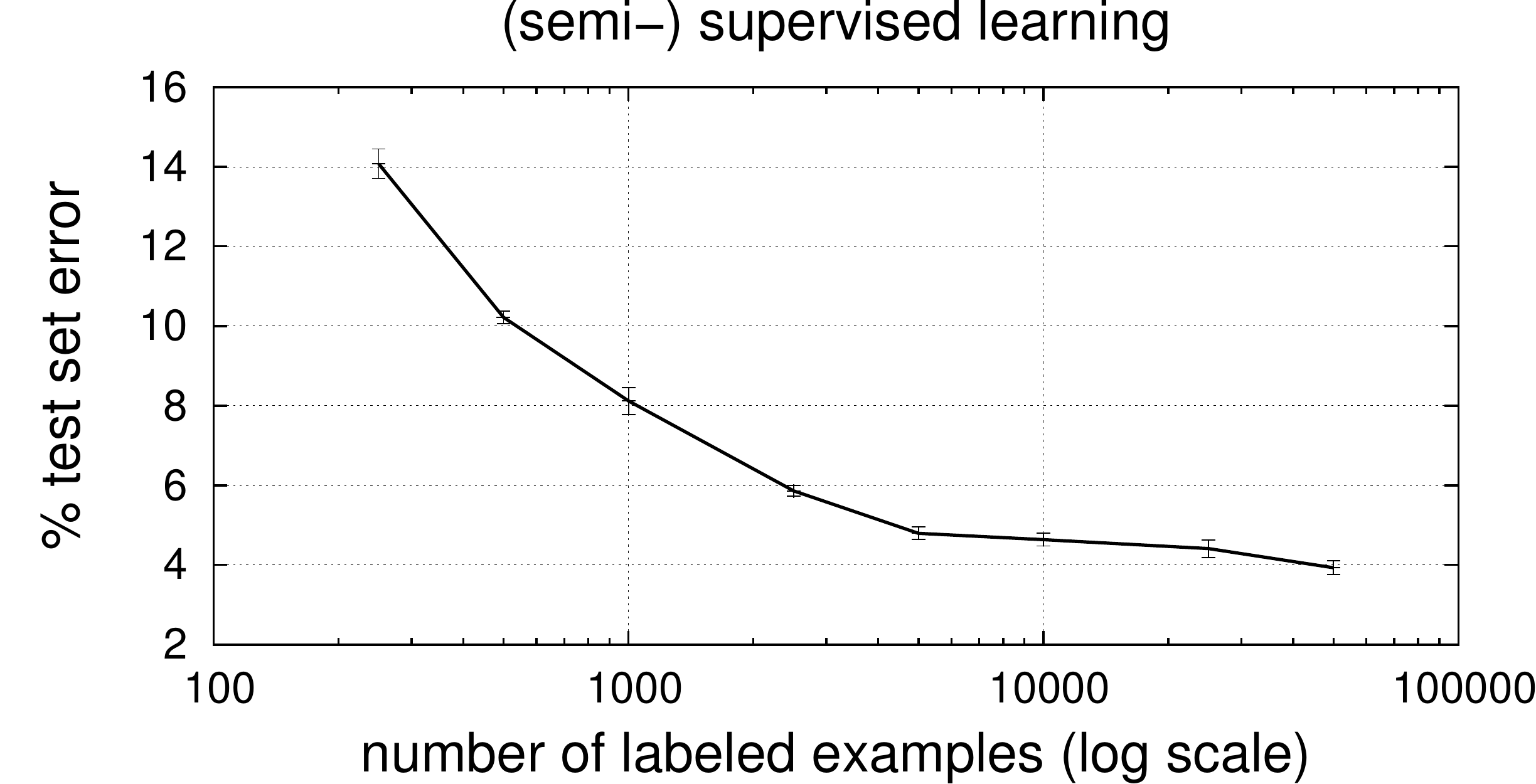}}
\caption{Results of the semi-supervised training runs. The x axis shows the number of training examples that were labeled (of 50,000 total), the y axis shows the resulting error rate on the 10,000 example test set.}
\label{results_labeled}
\end{center}
\vskip -0.2in
\end{figure}

\begin{figure}[ht]
\vskip 0.2in
\begin{center}
\centerline{\includegraphics[width=\columnwidth]{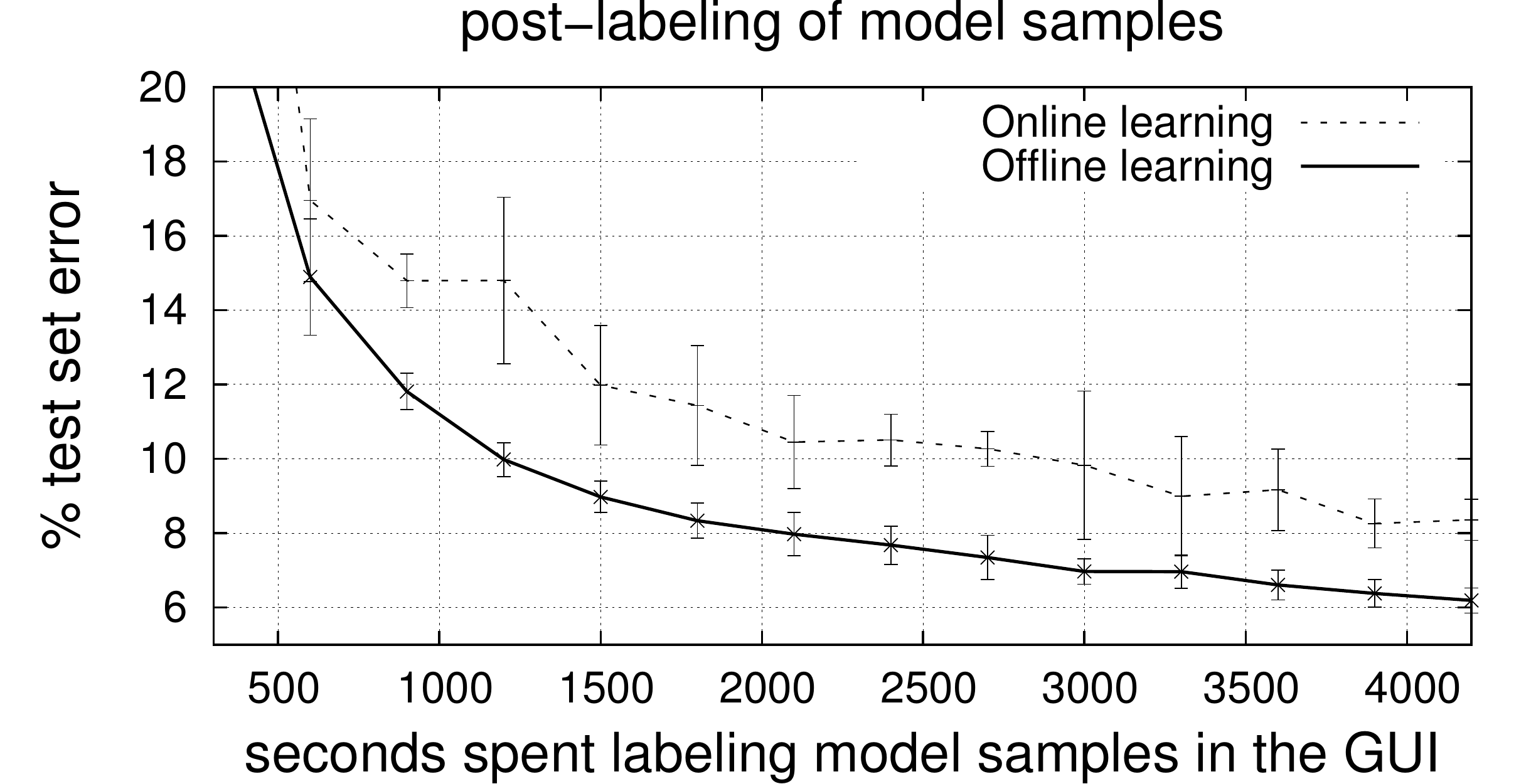}}
\caption{Results of the post-labeling training runs. The x axis shows the number of seconds spent on labeling samples from the model, the y axis shows the resulting error rate on the 10.000 example test set. Offline training yields better results than online learning.}
\label{results_postlabeling_on_and_offline}
\end{center}
\vskip -0.2in
\end{figure}

We test both the RBM trained with the standard (semi-) supervised approach as well as the post-labeling RBM using the MNIST test set with 10,000 labeled images.\\
Figure \ref{results_labeled} shows the resulting test set error rate of the RBM trained using the standard approach. Having only 500 of 50,000 images labeled results in a classification error of approx 14\%. On increasing the number of labeled images, the error rate drops quickly and reaches its minimum of approx. 4\% on a fully labeled training set.\\
Figure \ref{results_postlabeling_on_and_offline} shows the test set error of the RBM trained using the post-labeling approach. Both online learning and offline learning results show high initial error rates and a fast drop on increasing GUI time. However, the classification error of epoch-wise offline learning is constantly smaller. It reaches a performance of around 6.2\% error after 4200 seconds of labeling model samples.\\
Although our goal is to compare (semi-) supervised and post-labeling approach, we do not plot the results in a single figure because they do not share a common x axis. In order to compare the results, we have to make an estimation on the time required to label static images. Test showed that 1.5-2 seconds per labeled image is a realistic labeling rate. Given this labeling rate, the standard and the post-labeling approach show similar error rates given the labeling time. When spending 2,000 seconds on labeling, both approaches show a test set error around 8\% . Accordingly, the error rates for 4,000 seconds labeling time are around 6.5\% .  \\
\subsection{Biases to the results}
\label{biases}
The results shown above are biased in two ways. First, our  initial parameter choice for the the unsupervised model was influenced by our background knowledge from previous supervised tests with the MNIST data set. On a genuinely new training set, we wouldn't possess such knowledge and would have to rely on the reconstruction error only (see section \ref{models}). On the other hand, our results are biased by the fact that we use the labels of the official test set, which almost certainly come from a different distribution than the ones given by our labelers during training (consider the ambiguity of sevens and ones or fours and nines, given the cultural background). If all labels (test and training) origin from the same distribution, the test error rate will most probably be lower. The displayed results of the supervised model can profit from this fact, as opposed to the results of the  post-labeling model.\\
It is not known whether the MNIST labels were double-checked in order to get error-free labels (at least for the non-ambiguous cases). If there is more than one labeling pass, the required time increases accordingly in the standard approach.
\section{Discussion}
\label{discussion}
The results show that the post-labeling approach is, in gereral, competitive to the standard approach in terms of the resulting classification quality. It is likely that, by following the low-energy ravines, the model displays samples that resemble a class, but are not part of the training data. These samples can then be labeled by the GUI user.\\  
On the other hand, the post-labeling approach has a number of drawbacks. As mentioned above, the initial unsupervised training must rely on metrics such as the reconstruction error. Also, the quality of the labeled model samples is not as high as the quality of labeled real-world examples. As the displayed image is constantly changing, there are almost certainly some mislabeled or low-quality samples. Nevertheless it should be possible to use the labeled samples as a whole to train the label weights of a different model than the one they originated from, as most of them genuineley represent the classes.\\
Another drawback of this approach is that it is crucial to have meaningful reconstructions of the original input. They have to be clearly distinguishable from one another by a human observer, and more or less stable on repeated Gibbs sampling. Especially when dealing with real-world (and thus real-valued) images, this sets a high standard for the unsupervised model. The approach is, however, independent of the model type and can, e.g., be used with higher-order Boltzmann Machines to model covariances in the dataset to better model real-world images \cite{Ranzato2010}.\\
The approach can, in principle, be combined with classical semi-supervised learning, e.g. by initializing the label learning procedure with some labeled images in the training set or to get a better understanding of parameter settings by using a small labeled validation set.
\section{Conclusion and future work}
We proposed a different approach for training a classification model. Using the MNIST set of handwritten digits, we showed that it is feasible to train an RBM on unlabeled data first and subsequently label model samples using a GUI. This approach presents an alternative to semi-supervised learning, but does not reach the classification performance of a model trained on fully labeled data given the tested labeling times. An interesting question for further research is whether it is possible to also improve the model quality with respect to the data using the post-labeling GUI. That is, to capture user input during the interactive learning phase (such as "I see only noise") to improve the quality of the weights $w_{ij}$ connecting the visible and the hidden neurons.


\begin{thebibliography}{16}
\providecommand{\natexlab}[1]{#1}
\providecommand{\url}[1]{\texttt{#1}}
\expandafter\ifx\csname urlstyle\endcsname\relax
  \providecommand{\doi}[1]{doi: #1}\else
  \providecommand{\doi}{doi: \begingroup \urlstyle{rm}\Url}\fi

\bibitem[Bengio et~al.(2006)Bengio, Lamblin, Popovici, and
  Larochelle]{BengioTR2006}
Yoshua Bengio, Pascal Lamblin, Dan Popovici, and Hugo Larochelle.
\newblock Greedy layer-wise training of deep networks.
\newblock Technical Report 1282, Dept. IRO, Université de Montréal, August
  2006.

\bibitem[Chapelle et~al.(2010)Chapelle, Schoelkopf, and Zien]{Chapelle2010}
Olivier Chapelle, Bernhard Schoelkopf, and Alexander Zien.
\newblock \emph{Semi-Supervised Learning}.
\newblock The MIT Press, 1st edition, 2010.
\newblock ISBN 0262514125, 9780262514125.

\bibitem[Cohn et~al.(1994)Cohn, Atlas, and Ladner]{Cohn1994}
David Cohn, Les Atlas, and Richard Ladner.
\newblock Improving generalization with active learning.
\newblock \emph{Machine Learning}, 15:\penalty0 201--221, 1994.

\bibitem[G.~E.~Dahl and E.Hinton(2010)]{Dahl2010}
A.~Mohamed G.~E.~Dahl, M~Ranzato and G.~E.Hinton.
\newblock Phone recognition with the mean-covariance restricted boltzmann
  machine.
\newblock \emph{Advances in Neural Information Processing}, 23, 2010.

\bibitem[Hinton et~al.(2006)Hinton, Osindero, and Teh]{Hinton2006}
G.~E. Hinton, S.~Osindero, and Y.~Teh.
\newblock A fast learning algorithm for deep belief nets.
\newblock \emph{Neural Computation}, 18:\penalty0 1527--1554, 2006.

\bibitem[Hinton(2010)]{Hinton2010techreport}
Geoffrey Hinton.
\newblock A practical guide to training restricted boltzmann machines.
\newblock Technical Report UTML TR 2010–003, Department of Computer Science,
  University of Toronto, 2010.

\bibitem[Hinton and Salakhutdinov(2011)]{HintonSalakhutdinov2011}
Geoffrey Hinton and Ruslan Salakhutdinov.
\newblock Discovering binary codes for documents by learning deep generative
  models.
\newblock \emph{Topics in Cognitive Science}, 3\penalty0 (1):\penalty0 74--91,
  2011.
\newblock ISSN 1756-8765.
\newblock \doi{10.1111/j.1756-8765.2010.01109.x}.
\newblock URL \url{http://dx.doi.org/10.1111/j.1756-8765.2010.01109.x}.

\bibitem[Hinton(2002)]{Hinton2002}
Geoffrey~E. Hinton.
\newblock Training products of experts by minimizing contrastive divergence.
\newblock \emph{Neural Computation}, 14:\penalty0 1771–1800, 2002.

\bibitem[LeCun and Cortes()]{MNIST}
Courant Institute~NYU LeCun, Y. and Google Labs New~York Cortes, C.
\newblock The mnist database of handwritten digits.

\bibitem[Ranzato and Hinton(2010)]{Ranzato2010}
M.~Ranzato and G.~E. Hinton.
\newblock Modeling pixel means and covariances using factored third-order
  boltzmann machines.
\newblock In \emph{IEEE Conference on Computer Vision and Pattern Recognition},
  2010.

\bibitem[Smolensky(1986)]{Smolensky1986}
P.~Smolensky.
\newblock \emph{Parallel distributed processing: explorations in the
  microstructure of cognition}, volume~1, chapter Information processing in
  dynamical systems: foundations of harmony theory, pages 194--281.
\newblock MIT Press, Cambridge, MA, USA, 1986.

\bibitem[Tian et~al.(2010)Tian, Liu, Xiao, Wen, and Tang]{Tian2010}
Yuandong Tian, Wei Liu, Rong Xiao, Fang Wen, and Xiaoou Tang.
\newblock A face annotation framework with partial clustering and interactive
  labeling.
\newblock In \emph{IEEE Conference on Computer Vision and Pattern Recognition},
  2010.

\bibitem[Tieleman(2008)]{Tieleman2008}
Tijmen Tieleman.
\newblock Training restricted boltzmann machines using approximations to the
  likelihood gradient.
\newblock In \emph{Proceedings of the 25th international conference on Machine
  learning}, ICML '08, pages 1064--1071, New York, NY, USA, 2008. ACM.
\newblock ISBN 978-1-60558-205-4.
\newblock \doi{http://doi.acm.org/10.1145/1390156.1390290}.
\newblock URL \url{http://doi.acm.org/10.1145/1390156.1390290}.

\bibitem[Tieleman and Hinton(2009)]{Tieleman2009}
Tijmen Tieleman and Geoffrey Hinton.
\newblock Using fast weights to improve persistent contrastive divergence.
\newblock In \emph{Proceedings of the 26th Annual International Conference on
  Machine Learning}, ICML '09, pages 1033--1040, New York, NY, USA, 2009. ACM.
\newblock ISBN 978-1-60558-516-1.
\newblock \doi{http://doi.acm.org/10.1145/1553374.1553506}.
\newblock URL \url{http://doi.acm.org/10.1145/1553374.1553506}.

\bibitem[Welling et~al.(2005)Welling, Rosen-Zvi, and \&~Hinton]{Welling2005}
M.~Welling, M.~Rosen-Zvi, and G.~\&~Hinton.
\newblock Exponential family harmoniums with an application to information
  retrieval.
\newblock \emph{Advances in Neural Information Processing Systems}, 17, 2005.

\bibitem[Zhu and Goldberg(2009)]{zhu2009}
Xiaojin Zhu and Andrew~B. Goldberg.
\newblock \emph{Introduction to Semi-Supervised Learning Synthesis Lectures on
  Artificial Intelligence and Machine Learning}.
\newblock Morgan \& Claypool, 2009.

\end{thebibliography}
\end{document}